\documentclass[conference]{IEEEtran}
\usepackage[T1]{fontenc}
\usepackage{amsmath,amssymb,amsthm,booktabs,graphicx,url,microtype,placeins}
\newtheorem{proposition}{Proposition}
\graphicspath{{figures/}}
\usepackage{tikz}
\definecolor{figblue}{HTML}{00629B}
\definecolor{figorange}{HTML}{E87722}
\definecolor{figgreen}{HTML}{00843D}
\definecolor{figgray}{HTML}{5B6770}
\definecolor{figlight}{HTML}{D9E2E9}
\definecolor{figink}{HTML}{202124}
\definecolor{fignavy}{HTML}{234E70}
\definecolor{figteal}{HTML}{5D8F91}
\definecolor{figslate}{HTML}{A8B4C0}
\expandafter\def\csname paperfigurearchitecture.pdf\endcsname{%
\begin{tikzpicture}[x=1in,y=1in,>=stealth]
\path[use as bounding box] (0,0) rectangle (7,2.05);
\node[anchor=west,align=center,inner sep=0pt,text=figink,font={\sffamily\fontsize{8.5}{10.2}\selectfont\bfseries}] at (0.08000,1.93000) {Compute once};
\path[draw=figblue,fill=figblue!6,line width=.8pt] (0.08000,1.22000) rectangle (1.43000,1.77000);
\node[anchor=center,align=center,inner sep=0pt,text=figink,font={\sffamily\fontsize{8}{9.6}\selectfont\bfseries}] at (0.75500,1.59000) {Known case cues};
\node[anchor=center,align=center,inner sep=0pt,text=figink,font={\sffamily\fontsize{7.5}{9.0}\selectfont}] at (0.75500,1.37000) {$x_j$ and $x_t$};
\path[draw=figblue,fill=figblue!6,line width=.8pt] (1.77000,1.22000) rectangle (3.39000,1.77000);
\node[anchor=center,align=center,inner sep=0pt,text=figink,font={\sffamily\fontsize{8}{9.6}\selectfont\bfseries}] at (2.58000,1.59000) {Encoder and routing};
\node[anchor=center,align=center,inner sep=0pt,text=figink,font={\sffamily\fontsize{7.5}{9.0}\selectfont}] at (2.58000,1.37000) {$b,\ A$ or $\phi;\ \alpha_{tj}$};
\path[draw=figblue,fill=figblue!6,line width=.8pt] (3.72000,1.22000) rectangle (5.12000,1.77000);
\node[anchor=center,align=center,inner sep=0pt,text=figink,font={\sffamily\fontsize{8}{9.6}\selectfont\bfseries}] at (4.42000,1.59000) {Full-history maps};
\node[anchor=center,align=center,inner sep=0pt,text=figink,font={\sffamily\fontsize{7.5}{9.0}\selectfont}] at (4.42000,1.37000) {$b_t,\ p_j,\ W_{tj}$};
\path[draw=figblue,fill=figblue!6,line width=.8pt] (5.46000,1.22000) rectangle (6.90000,1.77000);
\node[anchor=center,align=center,inner sep=0pt,text=figink,font={\sffamily\fontsize{8}{9.6}\selectfont\bfseries}] at (6.18000,1.59000) {Width order};
\node[anchor=center,align=center,inner sep=0pt,text=figink,font={\sffamily\fontsize{7.5}{9.0}\selectfont}] at (6.18000,1.37000) {Rank responses};
\draw[draw=figblue,line width=1pt,->] (1.46000,1.50000) -- (1.74000,1.50000);
\draw[draw=figblue,line width=1pt,->] (3.42000,1.50000) -- (3.69000,1.50000);
\draw[draw=figblue,line width=1pt,->] (5.15000,1.50000) -- (5.43000,1.50000);
\node[anchor=west,align=center,inner sep=0pt,text=figink,font={\sffamily\fontsize{8.5}{10.2}\selectfont\bfseries}] at (0.08000,0.90000) {Update after each retrieval};
\path[draw=figblue,fill=figblue!6,line width=.8pt] (0.08000,0.15000) rectangle (1.83000,0.70000);
\node[anchor=center,align=center,inner sep=0pt,text=figink,font={\sffamily\fontsize{8}{9.6}\selectfont\bfseries}] at (0.95500,0.52000) {Observed responses};
\node[anchor=center,align=center,inner sep=0pt,text=figink,font={\sffamily\fontsize{7.5}{9.0}\selectfont}] at (0.95500,0.30000) {$O$ observed; $U$ hidden};
\path[draw=figblue,fill=figblue!6,line width=.8pt] (2.26000,0.15000) rectangle (4.68000,0.70000);
\node[anchor=center,align=center,inner sep=0pt,text=figink,font={\sffamily\fontsize{8}{9.6}\selectfont\bfseries}] at (3.47000,0.52000) {Prediction and completion bounds};
\node[anchor=center,align=center,inner sep=0pt,text=figink,font={\sffamily\fontsize{7.5}{9.0}\selectfont}] at (3.47000,0.30000) {$\hat\ell_t;\quad L_{tab}=\min_{y_U}(\ell_{ta}-\ell_{tb})$};
\path[draw=figblue,fill=figblue!6,line width=.8pt] (5.10000,0.15000) rectangle (6.90000,0.70000);
\node[anchor=center,align=center,inner sep=0pt,text=figink,font={\sffamily\fontsize{8}{9.6}\selectfont\bfseries}] at (6.00000,0.52000) {Class and certificate};
\node[anchor=center,align=center,inner sep=0pt,text=figink,font={\sffamily\fontsize{7.5}{9.0}\selectfont}] at (6.00000,0.30000) {Certified or unresolved};
\draw[draw=figblue,line width=1pt,->] (1.86000,0.42000) -- (2.23000,0.42000);
\draw[draw=figblue,line width=1pt,->] (4.71000,0.42000) -- (5.07000,0.42000);
\draw[draw=figgray,line width=0.8pt,->] (4.42000,1.22000) -- (4.42000,0.73000);
\draw[draw=figgray,line width=0.7pt] (6.18000,1.22000) -- (6.18000,1.00000);
\draw[draw=figgray,line width=0.7pt] (6.18000,1.00000) -- (0.02000,1.00000);
\draw[draw=figgray,line width=0.7pt] (0.02000,1.00000) -- (0.02000,0.42000);
\draw[draw=figgray,line width=0.8pt,->] (0.02000,0.42000) -- (0.07500,0.42000);
\end{tikzpicture}
}
\expandafter\def\csname paperfigurerq_evidence.pdf\endcsname{%
\begin{tikzpicture}[x=1in,y=1in,>=stealth]
\path[use as bounding box] (0,0) rectangle (7,3.9);
\node[anchor=west,align=center,inner sep=0pt,text=figink,font={\sffamily\fontsize{8.5}{10.2}\selectfont\bfseries}] at (0.08000,3.76000) {A  Loss differences};
\draw[draw=figlight,line width=0.7pt] (2.28118,2.32000) -- (2.28118,3.55000);
\node[anchor=center,align=center,inner sep=0pt,text=figink,font={\sffamily\fontsize{7}{8.4}\selectfont}] at (2.28118,2.21000) {-0.14};
\draw[draw=figlight,line width=0.7pt] (3.40588,2.32000) -- (3.40588,3.55000);
\node[anchor=center,align=center,inner sep=0pt,text=figink,font={\sffamily\fontsize{7}{8.4}\selectfont}] at (3.40588,2.21000) {-0.10};
\draw[draw=figlight,line width=0.7pt] (4.53059,2.32000) -- (4.53059,3.55000);
\node[anchor=center,align=center,inner sep=0pt,text=figink,font={\sffamily\fontsize{7}{8.4}\selectfont}] at (4.53059,2.21000) {-0.06};
\draw[draw=figlight,line width=0.7pt] (5.65529,2.32000) -- (5.65529,3.55000);
\node[anchor=center,align=center,inner sep=0pt,text=figink,font={\sffamily\fontsize{7}{8.4}\selectfont}] at (5.65529,2.21000) {-0.02};
\draw[draw=figlight,line width=0.7pt] (6.21765,2.32000) -- (6.21765,3.55000);
\node[anchor=center,align=center,inner sep=0pt,text=figink,font={\sffamily\fontsize{7}{8.4}\selectfont}] at (6.21765,2.21000) {0.00};
\draw[draw=figlight,line width=0.7pt] (6.78000,2.32000) -- (6.78000,3.55000);
\node[anchor=center,align=center,inner sep=0pt,text=figink,font={\sffamily\fontsize{7}{8.4}\selectfont}] at (6.78000,2.21000) {0.02};
\draw[draw=figgray,line width=0.8pt,dashed] (6.21765,2.32000) -- (6.21765,3.55000);
\node[anchor=east,align=center,inner sep=0pt,text=figink,font={\sffamily\fontsize{7.5}{9.0}\selectfont}] at (1.86000,3.40000) {Ridge vs population};
\draw[draw=figblue,line width=1pt] (5.82050,3.46500) -- (6.13902,3.46500);
\draw[draw=figblue,line width=1pt] (5.82050,3.44000) -- (5.82050,3.49000);
\draw[draw=figblue,line width=1pt] (6.13902,3.44000) -- (6.13902,3.49000);
\fill[figblue] (5.97738,3.46500) circle[radius=0.027in];
\draw[draw=figorange,line width=1pt] (5.67361,3.40000) -- (6.02057,3.40000);
\draw[draw=figorange,line width=1pt] (5.67361,3.37500) -- (5.67361,3.42500);
\draw[draw=figorange,line width=1pt] (6.02057,3.37500) -- (6.02057,3.42500);
\path[draw=figorange,fill=figorange,line width=.8pt] (5.82687,3.37300) rectangle (5.88087,3.42700);
\draw[draw=figgreen,line width=1pt] (2.40649,3.33500) -- (3.03285,3.33500);
\draw[draw=figgreen,line width=1pt] (2.40649,3.31000) -- (2.40649,3.36000);
\draw[draw=figgreen,line width=1pt] (3.03285,3.31000) -- (3.03285,3.36000);
\fill[figgreen] (2.73119,3.36200) -- (2.70419,3.30800) -- (2.75819,3.30800) -- cycle;
\node[anchor=east,align=center,inner sep=0pt,text=figink,font={\sffamily\fontsize{7.5}{9.0}\selectfont}] at (1.86000,3.11000) {Ridge vs frequency};
\draw[draw=figblue,line width=1pt] (6.12099,3.17500) -- (6.54454,3.17500);
\draw[draw=figblue,line width=1pt] (6.12099,3.15000) -- (6.12099,3.20000);
\draw[draw=figblue,line width=1pt] (6.54454,3.15000) -- (6.54454,3.20000);
\fill[figblue] (6.32997,3.17500) circle[radius=0.027in];
\draw[draw=figorange,line width=1pt] (6.06426,3.11000) -- (6.66314,3.11000);
\draw[draw=figorange,line width=1pt] (6.06426,3.08500) -- (6.06426,3.13500);
\draw[draw=figorange,line width=1pt] (6.66314,3.08500) -- (6.66314,3.13500);
\path[draw=figorange,fill=figorange,line width=.8pt] (6.31581,3.08300) rectangle (6.36981,3.13700);
\draw[draw=figgreen,line width=1pt] (4.30343,3.04500) -- (5.42361,3.04500);
\draw[draw=figgreen,line width=1pt] (4.30343,3.02000) -- (4.30343,3.07000);
\draw[draw=figgreen,line width=1pt] (5.42361,3.02000) -- (5.42361,3.07000);
\fill[figgreen] (4.84472,3.07200) -- (4.81772,3.01800) -- (4.87172,3.01800) -- cycle;
\node[anchor=east,align=center,inner sep=0pt,text=figink,font={\sffamily\fontsize{7.5}{9.0}\selectfont}] at (1.86000,2.82000) {Ridge vs matched frequency};
\draw[draw=figblue,line width=1pt] (5.96446,2.88500) -- (6.24058,2.88500);
\draw[draw=figblue,line width=1pt] (5.96446,2.86000) -- (5.96446,2.91000);
\draw[draw=figblue,line width=1pt] (6.24058,2.86000) -- (6.24058,2.91000);
\fill[figblue] (6.09966,2.88500) circle[radius=0.027in];
\draw[draw=figorange,line width=1pt] (5.83558,2.82000) -- (6.13122,2.82000);
\draw[draw=figorange,line width=1pt] (5.83558,2.79500) -- (5.83558,2.84500);
\draw[draw=figorange,line width=1pt] (6.13122,2.79500) -- (6.13122,2.84500);
\path[draw=figorange,fill=figorange,line width=.8pt] (5.96294,2.79300) rectangle (6.01694,2.84700);
\draw[draw=figgreen,line width=1pt] (3.20498,2.75500) -- (3.71865,2.75500);
\draw[draw=figgreen,line width=1pt] (3.20498,2.73000) -- (3.20498,2.78000);
\draw[draw=figgreen,line width=1pt] (3.71865,2.73000) -- (3.71865,2.78000);
\fill[figgreen] (3.45751,2.78200) -- (3.43051,2.72800) -- (3.48451,2.72800) -- cycle;
\node[anchor=east,align=center,inner sep=0pt,text=figink,font={\sffamily\fontsize{7.5}{9.0}\selectfont}] at (1.86000,2.53000) {Width vs random};
\draw[draw=figblue,line width=1pt] (5.93763,2.59500) -- (6.21595,2.59500);
\draw[draw=figblue,line width=1pt] (5.93763,2.57000) -- (5.93763,2.62000);
\draw[draw=figblue,line width=1pt] (6.21595,2.57000) -- (6.21595,2.62000);
\fill[figblue] (6.08243,2.59500) circle[radius=0.027in];
\draw[draw=figorange,line width=1pt] (5.99276,2.53000) -- (6.23572,2.53000);
\draw[draw=figorange,line width=1pt] (5.99276,2.50500) -- (5.99276,2.55500);
\draw[draw=figorange,line width=1pt] (6.23572,2.50500) -- (6.23572,2.55500);
\path[draw=figorange,fill=figorange,line width=.8pt] (6.09154,2.50300) rectangle (6.14554,2.55700);
\draw[draw=figgreen,line width=1pt] (5.09310,2.46500) -- (5.40770,2.46500);
\draw[draw=figgreen,line width=1pt] (5.09310,2.44000) -- (5.09310,2.49000);
\draw[draw=figgreen,line width=1pt] (5.40770,2.44000) -- (5.40770,2.49000);
\fill[figgreen] (5.24830,2.49200) -- (5.22130,2.43800) -- (5.27530,2.43800) -- cycle;
\node[anchor=center,align=center,inner sep=0pt,text=figink,font={\sffamily\fontsize{8}{9.6}\selectfont}] at (4.40000,2.03000) {Paired loss difference (nats); negative favors first method};
\fill[figblue] (2.30000,3.75000) circle[radius=0.028in];
\node[anchor=west,align=center,inner sep=0pt,text=figink,font={\sffamily\fontsize{7.5}{9.0}\selectfont}] at (2.40000,3.75000) {RepVig};
\path[draw=figorange,fill=figorange,line width=.8pt] (3.73200,3.72200) rectangle (3.78800,3.77800);
\node[anchor=west,align=center,inner sep=0pt,text=figink,font={\sffamily\fontsize{7.5}{9.0}\selectfont}] at (3.86000,3.75000) {Schmieding};
\fill[figgreen] (5.22000,3.77800) -- (5.19200,3.72200) -- (5.24800,3.72200) -- cycle;
\node[anchor=west,align=center,inner sep=0pt,text=figink,font={\sffamily\fontsize{7.5}{9.0}\selectfont}] at (5.32000,3.75000) {Clinicians};
\node[anchor=west,align=center,inner sep=0pt,text=figink,font={\sffamily\fontsize{8.5}{10.2}\selectfont\bfseries}] at (0.10000,1.83000) {B  Coverage};
\draw[draw=figlight,line width=0.7pt] (0.52000,0.47000) -- (3.18000,0.47000);
\node[anchor=east,align=center,inner sep=0pt,text=figink,font={\sffamily\fontsize{7}{8.4}\selectfont}] at (0.44000,0.47000) {0};
\draw[draw=figlight,line width=0.7pt] (0.52000,0.74500) -- (3.18000,0.74500);
\node[anchor=east,align=center,inner sep=0pt,text=figink,font={\sffamily\fontsize{7}{8.4}\selectfont}] at (0.44000,0.74500) {25};
\draw[draw=figlight,line width=0.7pt] (0.52000,1.02000) -- (3.18000,1.02000);
\node[anchor=east,align=center,inner sep=0pt,text=figink,font={\sffamily\fontsize{7}{8.4}\selectfont}] at (0.44000,1.02000) {50};
\draw[draw=figlight,line width=0.7pt] (0.52000,1.29500) -- (3.18000,1.29500);
\node[anchor=east,align=center,inner sep=0pt,text=figink,font={\sffamily\fontsize{7}{8.4}\selectfont}] at (0.44000,1.29500) {75};
\draw[draw=figlight,line width=0.7pt] (0.52000,1.57000) -- (3.18000,1.57000);
\node[anchor=east,align=center,inner sep=0pt,text=figink,font={\sffamily\fontsize{7}{8.4}\selectfont}] at (0.44000,1.57000) {100};
\path[draw=figslate,fill=figslate,line width=.8pt] (0.52000,1.66000) rectangle (0.60500,1.71000);
\node[anchor=west,align=center,inner sep=0pt,text=figink,font={\sffamily\fontsize{7}{8.4}\selectfont}] at (0.64000,1.68500) {Zero};
\path[draw=figslate,fill=figslate,line width=.8pt] (0.55000,0.47000) rectangle (0.73000,0.93481);
\path[draw=figslate,fill=figslate,line width=.8pt] (1.44000,0.47000) rectangle (1.62000,0.84035);
\path[draw=figslate,fill=figslate,line width=.8pt] (2.33000,0.47000) rectangle (2.51000,0.50263);
\path[draw=fignavy,fill=fignavy,line width=.8pt] (1.41000,1.66000) rectangle (1.49500,1.71000);
\node[anchor=west,align=center,inner sep=0pt,text=figink,font={\sffamily\fontsize{7}{8.4}\selectfont}] at (1.53000,1.68500) {Half: all};
\path[draw=fignavy,fill=fignavy,line width=.8pt] (0.74000,0.47000) rectangle (0.92000,1.36426);
\path[draw=fignavy,fill=fignavy,line width=.8pt] (1.63000,0.47000) rectangle (1.81000,1.33122);
\path[draw=fignavy,fill=fignavy,line width=.8pt] (2.52000,0.47000) rectangle (2.70000,0.99901);
\path[draw=figteal,fill=figteal,line width=.8pt] (2.30000,1.66000) rectangle (2.38500,1.71000);
\node[anchor=west,align=center,inner sep=0pt,text=figink,font={\sffamily\fontsize{7}{8.4}\selectfont}] at (2.42000,1.68500) {Half: active};
\path[draw=figteal,fill=figteal,line width=.8pt] (0.93000,0.47000) rectangle (1.11000,0.79107);
\path[draw=figteal,fill=figteal,line width=.8pt] (1.82000,0.47000) rectangle (2.00000,0.61934);
\path[draw=figteal,fill=figteal,line width=.8pt] (2.71000,0.47000) rectangle (2.89000,0.72727);
\node[anchor=center,align=center,inner sep=0pt,text=figink,font={\sffamily\fontsize{7.4}{8.88}\selectfont}] at (0.83000,0.32000) {RepVig};
\node[anchor=center,align=center,inner sep=0pt,text=figink,font={\sffamily\fontsize{7.4}{8.88}\selectfont}] at (1.72000,0.32000) {Schmieding};
\node[anchor=center,align=center,inner sep=0pt,text=figink,font={\sffamily\fontsize{7.4}{8.88}\selectfont}] at (2.61000,0.32000) {Clinicians};
\node[anchor=center,align=center,inner sep=0pt,text=figink,font={\sffamily\fontsize{8}{9.6}\selectfont}] at (1.85000,0.09000) {Certified targets (\%)};
\node[anchor=west,align=center,inner sep=0pt,text=figink,font={\sffamily\fontsize{8.5}{10.2}\selectfont\bfseries}] at (3.63000,1.83000) {C  Prediction error};
\draw[draw=figlight,line width=0.7pt] (4.05000,0.47000) -- (6.71000,0.47000);
\node[anchor=east,align=center,inner sep=0pt,text=figink,font={\sffamily\fontsize{7}{8.4}\selectfont}] at (3.97000,0.47000) {0};
\draw[draw=figlight,line width=0.7pt] (4.05000,0.83667) -- (6.71000,0.83667);
\node[anchor=east,align=center,inner sep=0pt,text=figink,font={\sffamily\fontsize{7}{8.4}\selectfont}] at (3.97000,0.83667) {20};
\draw[draw=figlight,line width=0.7pt] (4.05000,1.20333) -- (6.71000,1.20333);
\node[anchor=east,align=center,inner sep=0pt,text=figink,font={\sffamily\fontsize{7}{8.4}\selectfont}] at (3.97000,1.20333) {40};
\draw[draw=figlight,line width=0.7pt] (4.05000,1.57000) -- (6.71000,1.57000);
\node[anchor=east,align=center,inner sep=0pt,text=figink,font={\sffamily\fontsize{7}{8.4}\selectfont}] at (3.97000,1.57000) {60};
\path[draw=fignavy,fill=fignavy,line width=.8pt] (4.05000,1.66000) rectangle (4.13500,1.71000);
\node[anchor=west,align=center,inner sep=0pt,text=figink,font={\sffamily\fontsize{7}{8.4}\selectfont}] at (4.17000,1.68500) {Certified};
\path[draw=fignavy,fill=fignavy,line width=.8pt] (4.08000,0.47000) rectangle (4.26000,0.93849);
\path[draw=fignavy,fill=fignavy,line width=.8pt] (4.97000,0.47000) rectangle (5.15000,0.88619);
\path[draw=fignavy,fill=fignavy,line width=.8pt] (5.86000,0.47000) rectangle (6.04000,0.70175);
\path[draw=figteal,fill=figteal,line width=.8pt] (4.94000,1.66000) rectangle (5.02500,1.71000);
\node[anchor=west,align=center,inner sep=0pt,text=figink,font={\sffamily\fontsize{7}{8.4}\selectfont}] at (5.06000,1.68500) {Confidence};
\path[draw=figteal,fill=figteal,line width=.8pt] (4.27000,0.47000) rectangle (4.45000,0.92027);
\path[draw=figteal,fill=figteal,line width=.8pt] (5.16000,0.47000) rectangle (5.34000,0.87202);
\path[draw=figteal,fill=figteal,line width=.8pt] (6.05000,0.47000) rectangle (6.23000,0.68116);
\path[draw=figslate,fill=figslate,line width=.8pt] (5.83000,1.66000) rectangle (5.91500,1.71000);
\node[anchor=west,align=center,inner sep=0pt,text=figink,font={\sffamily\fontsize{7}{8.4}\selectfont}] at (5.95000,1.68500) {Uncertified};
\path[draw=figslate,fill=figslate,line width=.8pt] (4.46000,0.47000) rectangle (4.64000,1.31158);
\path[draw=figslate,fill=figslate,line width=.8pt] (5.35000,0.47000) rectangle (5.53000,1.18558);
\path[draw=figslate,fill=figslate,line width=.8pt] (6.24000,0.47000) rectangle (6.42000,1.15032);
\node[anchor=center,align=center,inner sep=0pt,text=figink,font={\sffamily\fontsize{7.4}{8.88}\selectfont}] at (4.36000,0.32000) {RepVig};
\node[anchor=center,align=center,inner sep=0pt,text=figink,font={\sffamily\fontsize{7.4}{8.88}\selectfont}] at (5.25000,0.32000) {Schmieding};
\node[anchor=center,align=center,inner sep=0pt,text=figink,font={\sffamily\fontsize{7.4}{8.88}\selectfont}] at (6.14000,0.32000) {Clinicians};
\node[anchor=center,align=center,inner sep=0pt,text=figink,font={\sffamily\fontsize{8}{9.6}\selectfont}] at (5.38000,0.09000) {Response error (\%)};
\end{tikzpicture}
}
\expandafter\def\csname paperfigureworked_clinician.pdf\endcsname{%
\begin{tikzpicture}[x=1in,y=1in,>=stealth]
\path[use as bounding box] (0,0) rectangle (3.45,1.98);
\node[anchor=west,align=center,inner sep=0pt,text=figink,font={\sffamily\fontsize{8.5}{10.2}\selectfont\bfseries}] at (0.06000,1.88000) {Held-out clinician};
\node[anchor=west,align=center,inner sep=0pt,text=figink,font={\sffamily\fontsize{7.6}{9.12}\selectfont}] at (0.06000,1.66000) {Retrieve item 78: "absent"};
\draw[draw=figlight,line width=0.7pt] (0.88000,0.50000) -- (0.88000,1.44000);
\node[anchor=center,align=center,inner sep=0pt,text=figink,font={\sffamily\fontsize{7.5}{9.0}\selectfont}] at (0.88000,0.35000) {-1.5};
\draw[draw=figlight,line width=0.7pt] (1.49250,0.50000) -- (1.49250,1.44000);
\node[anchor=center,align=center,inner sep=0pt,text=figink,font={\sffamily\fontsize{7.5}{9.0}\selectfont}] at (1.49250,0.35000) {-1};
\draw[draw=figlight,line width=0.7pt] (2.10500,0.50000) -- (2.10500,1.44000);
\node[anchor=center,align=center,inner sep=0pt,text=figink,font={\sffamily\fontsize{7.5}{9.0}\selectfont}] at (2.10500,0.35000) {-0.5};
\draw[draw=figlight,line width=0.7pt] (2.71750,0.50000) -- (2.71750,1.44000);
\node[anchor=center,align=center,inner sep=0pt,text=figink,font={\sffamily\fontsize{7.5}{9.0}\selectfont}] at (2.71750,0.35000) {0};
\draw[draw=figlight,line width=0.7pt] (3.33000,0.50000) -- (3.33000,1.44000);
\node[anchor=center,align=center,inner sep=0pt,text=figink,font={\sffamily\fontsize{7.5}{9.0}\selectfont}] at (3.33000,0.35000) {0.5};
\draw[draw=figgray,line width=0.9pt,dashed] (2.71750,0.50000) -- (2.71750,1.44000);
\node[anchor=west,align=center,inner sep=0pt,text=figink,font={\sffamily\fontsize{8}{9.6}\selectfont\bfseries}] at (0.06000,1.32000) {Item 74};
\node[anchor=west,align=center,inner sep=0pt,text=figink,font={\sffamily\fontsize{7}{8.4}\selectfont}] at (0.06000,1.17000) {Newly certified};
\draw[draw=figlight,line width=1.3pt] (2.70687,1.32500) -- (2.76767,1.23500);
\path[draw=figgray,fill=figgray,line width=.8pt] (2.67687,1.29500) rectangle (2.73687,1.35500);
\fill[figblue] (2.76767,1.23500) circle[radius=0.03in];
\node[anchor=center,align=center,inner sep=0pt,text=figink,font={\sffamily\fontsize{7.1}{8.52}\selectfont}] at (2.73727,1.47000) {-0.009 to 0.041};
\node[anchor=west,align=center,inner sep=0pt,text=figink,font={\sffamily\fontsize{8}{9.6}\selectfont\bfseries}] at (0.06000,0.83000) {Item 25};
\node[anchor=west,align=center,inner sep=0pt,text=figink,font={\sffamily\fontsize{7}{8.4}\selectfont}] at (0.06000,0.68000) {Unresolved};
\draw[draw=figlight,line width=1.3pt] (1.33739,0.83500) -- (1.54132,0.74500);
\path[draw=figgray,fill=figgray,line width=.8pt] (1.30739,0.80500) rectangle (1.36739,0.86500);
\fill[figblue] (1.54132,0.74500) circle[radius=0.03in];
\node[anchor=center,align=center,inner sep=0pt,text=figink,font={\sffamily\fontsize{7.1}{8.52}\selectfont}] at (1.43936,0.98000) {-1.127 to -0.960};
\node[anchor=center,align=center,inner sep=0pt,text=figink,font={\sffamily\fontsize{7.5}{9.0}\selectfont}] at (2.03000,0.13000) {Minimum logit margin};
\path[draw=figgray,fill=figgray,line width=.8pt] (0.11500,0.29500) rectangle (0.16500,0.34500);
\node[anchor=west,align=center,inner sep=0pt,text=figink,font={\sffamily\fontsize{7}{8.4}\selectfont}] at (0.23000,0.32000) {25 ratings};
\fill[figblue] (0.14000,0.12000) circle[radius=0.025in];
\node[anchor=west,align=center,inner sep=0pt,text=figink,font={\sffamily\fontsize{7}{8.4}\selectfont}] at (0.23000,0.12000) {26 ratings};
\end{tikzpicture}
}
\newcommand{\paperfigure}[2]{\resizebox{#1}{!}{\csname paperfigure#2\endcsname}}
\usepackage{atbegshi}
\newcount\PaperPageCount
\AtBeginShipout{\global\advance\PaperPageCount by 1\relax
\ifnum\PaperPageCount>8\relax
\PackageError{PaperLength}{Manuscript exceeds eight pages}{Shorten the manuscript without reducing the standard font size.}\fi}
\title{Can the Remaining Answers Change the Prediction?\\Worst-Case Stability Checks\\for Personalized Rating Models}
\author{
\IEEEauthorblockN{
Priyanka Siddappa, Muntaqim Ahmed Raju,
Shifat Haider Al Amin, and Ruizhe Ma
}
\IEEEauthorblockA{
University of Massachusetts Lowell, Lowell, MA 01854, USA\\
Email: \{Priyanka\_Siddappa, MuntaqimAhmed\_Raju, 
MdShifat\_AlAmin, Ruizhe\_Ma\}@uml.edu
}
}
\begin{document}
\maketitle

\begin{abstract}
Missing ratings can overturn a personalized model's current prediction. For completed models with additive response scores, exact worst case bounds check all missing assignments without changing the current predictor. We evaluate this audit retrospectively in two self-triage cohorts, 108 clinicians rating three simulated psychiatric interviews, and 126 counseling demonstrations annotated by ten experts. Under shared random half histories, Fisher adaptation based on observed ratings reduces log loss relative to response frequency by 0.1071 nats in the clinician audit (95\% CI [0.0867,0.1261]) and 0.0223 in counseling (95\% CI [0.0121,0.0321]). The counseling test excludes target experts and source videos from training. Certification covers 5.4\% and 10.5\% of targets; fixed width acquisition raises coverage to 13.7\% and 28.6\% at the same budgets. With weights and responses held fixed, the observed-rating update also has lower loss than an affine partial update while certifying exactly the same evaluated targets. The certificate establishes agreement with the completed model, not correctness or stability at intermediate updates. These rating tasks test the method; clinical usefulness remains untested.

\end{abstract}

\begin{IEEEkeywords}
model auditing, trustworthy AI, clinical decision support, missing responses, completion certification
\end{IEEEkeywords}

\section{Introduction}
Personalized rating models use an individual's observed responses to predict how that person will rate other items. We study this setting in clinical and health-related rating tasks, including clinicians' symptom ratings of simulated psychiatric interviews \cite{lenz, multimodalreview,chexfuse,cirrguide}. When part of the calibration history is unavailable, the model must predict from incomplete information. Even a confident prediction may change once the remaining responses become available.

This raises a practical question: can any assignment of the missing ratings make the completed model disagree with the current class? The idea resembles calling an election when the leader wins under every allocation of the uncounted votes. Here, the uncounted inputs are missing ratings. A completion certificate establishes that the current class wins under every completed history.

\begin{figure*}[!t]
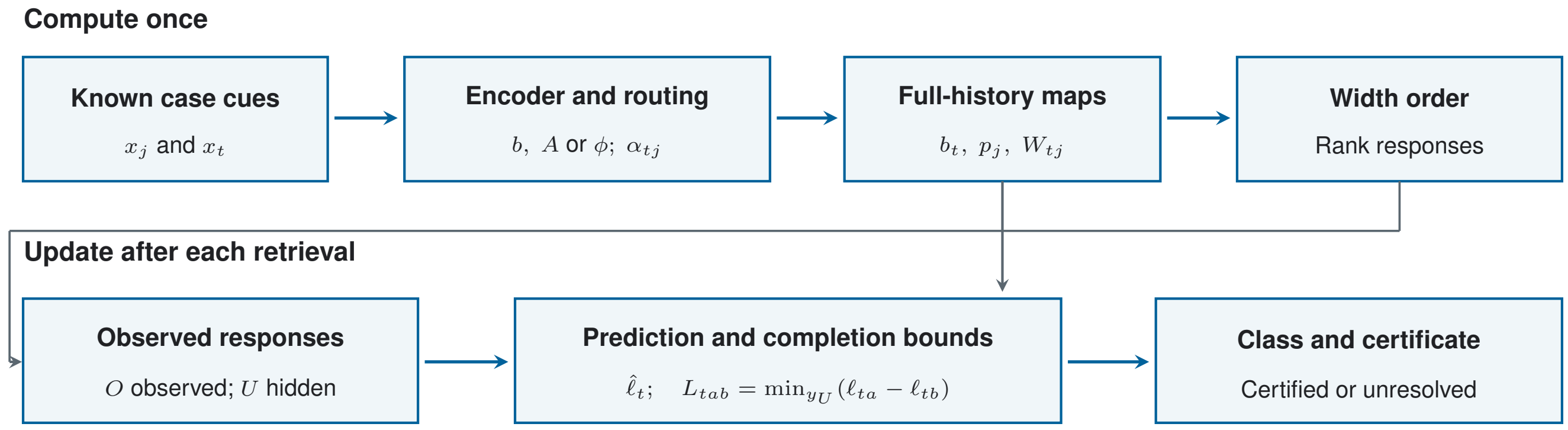

\centering\paperfigure{.98\textwidth}{architecture.pdf}
\caption{Completion auditing for a fixed model. Partial predictions may use observed curvature; optional width acquisition ranks responses once.}
\label{fig:architecture}
\end{figure*}

Our main contribution is to separate the prediction made from partial history from the calculation of its completion certificate (Fig.~\ref{fig:architecture}). When the completed model adds contributions from individual responses, separability gives exact bounds over all missing assignments. The partial predictor can still use a different update, such as Observed Fisher's update based on observed curvature. This allows us to audit its current class without changing its predictions. The guarantee concerns the fitted model's behavior; it does not establish correctness or provide a causal explanation.

We evaluate completion agreement, error, responsiveness to personal history, and fitted-model variation (RQ1); predictive gains across people, interviews, and categories (RQ2); and acquisition and stopping policies (RQ3). Models are fitted separately within four rating datasets, used as methodological testbeds.

\section{Related Work}

Prior certification studies ask whether predictions agree across plausible training datasets \cite{multiplicity,labelcert}, including missing-data completions for Naive Bayes \cite{nbmissing} and repairs for nearest-neighbor classifiers \cite{knnrepairs,alertconcordance}. Our work applies this perspective to incomplete personal histories, with the learned encoder and adaptation rule held fixed. The methodological distinction is the separation of the partial predictor from its completion audit: the current prediction can use observed curvature, while certification uses an affine map of the completed history.

Efficient adaptation provides the computational basis for this separation. Differentiable closed-form solvers combine learned representations with ridge regression \cite{r2d2,ttr,faast}, and ALPaCA learns features and priors for Bayesian regression \cite{alpaca}. Attentive neural processes instead aggregate context separately for each target \cite{anp}. These methods motivate our adaptation and attention studies. For certification, the relevant restriction is additive dependence on response labels; the cue encoder can remain nonlinear.

The resulting guarantee differs from those studied in selective classification and conformal prediction. Selective classification controls the tradeoff between accepted predictions and error \cite{selective,selectivednn}, while conformal methods address outcome coverage under missingness \cite{conformalmissing}. Completion agreement concerns the behavior of the fitted model, so it can certify an incorrect prediction. Our matched-coverage confidence comparisons assess this distinction empirically.

Adaptive testing connects completion auditing to stopping. Besides estimating latent traits \cite{cat}, projection-based rules examine whether hypothetical remaining answers can change a final classification \cite{catstop}. We instead certify individual predicted ratings over a fixed calibration history and distinguish stopping for one target from stopping for all targets sharing that history. EDDI and DIME guide acquisition through information gain \cite{eddi,dime,brig,afabench}. Our acquisition experiments examine a different tradeoff: whether selecting responses for lower predictive loss also improves completion coverage. They compare these objectives without claiming to reproduce the complete EDDI or DIME systems.

\section{Completion Audits for Neural Adaptation}
\paragraph{Overview} Calibration ratings describe one person's history; target ratings are what we predict. Item cues are known, but some calibration ratings are hidden. The audit checks whether the current class wins under every missing assignment. Coverage is the fraction of targets that pass.

\begin{figure}[!t]
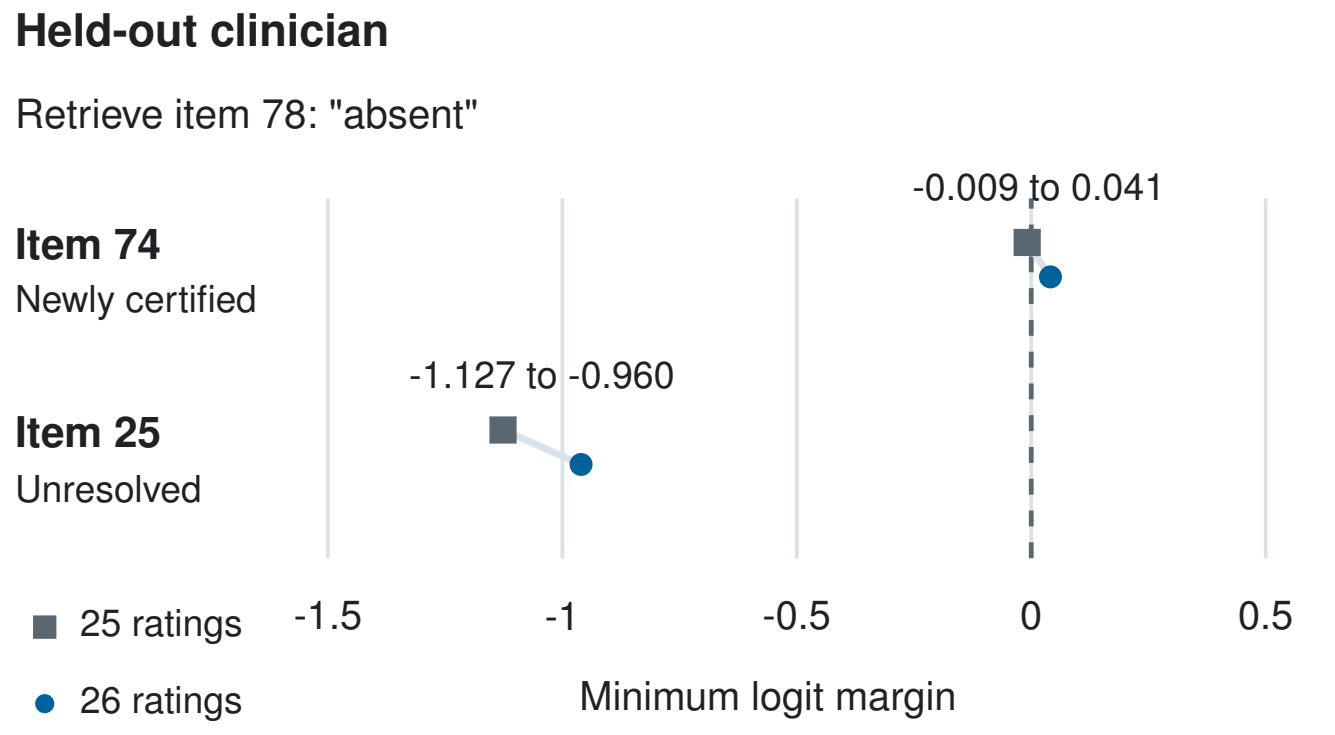

\centering\paperfigure{\columnwidth}{worked_clinician.pdf}
\caption{Minimum completion margins after 25 and 26 ratings. Retrieving item 78 certifies item 74; item 25 remains unresolved.}
\label{fig:example}
\end{figure}

For example, after 25 ratings, ridge predicts ``absent'' for item 74, but a missing-history completion can overturn it. Revealing item 78, rated ``absent,'' raises its worst case margin from $-.0087$ to $.0410$ and certifies the class. Item 25 remains unresolved (Fig.~\ref{fig:example}).

For one person, let $J=\{1,\ldots,H\}$ index calibration cases with known cues $x_j$ and responses $y_j\in\{1,\ldots,K\}$. Responses in $O\subseteq J$ are observed; those in $U=J\setminus O$ are hidden. All $T$ target cues $x_t$ are known before acquisition. Parameters are learned from other people. A completion assigns a category to each hidden response, allowing every assignment in $\{1,\ldots,K\}^{|U|}$. This set imposes no statistical independence assumption. Retrieval reveals a recorded response; cues and fitted maps stay fixed.

\subsection{Affine neural adapters}
An affine response map adds one contribution per rating to a baseline score. Its case encoder can be nonlinear; additivity is required only in the responses.
A nonlinear encoder produces population logits $b(x)\in\mathbb R^K$ and a basis $A(x)\in\mathbb R^{K\times r}$ centered over output classes. Define $p_j=\operatorname{softmax}(b_j)$, $S_j=\operatorname{diag}(p_j)-p_jp_j^\top$, and the one-hot vector $e_{y_j}$. All unqualified sums below run over $J$. With positive cue weights $\alpha_{tj}$ and damping $\lambda>0$, the Fisher adapter is
\begin{align}
D_t &= \lambda I+\sum_j\alpha_{tj}A_j^\top S_jA_j,\\
W_{tj}&=\alpha_{tj}A_tD_t^{-1}A_j^\top,\\
\ell_t(y)&=b_t+\sum_jW_{tj}(e_{y_j}-p_j). \label{eq:affine}
\end{align}
Global adaptation sets $\alpha_{tj}=1$. Routed adaptation uses $H\operatorname{softmax}_j(q(x_t)^\top k(x_j)/\sqrt r)$, where $q$ and $k$ are learned linear maps from cues to $\mathbb R^r$. The maps never see responses. Equation~\eqref{eq:affine} is one damped Fisher update from a zero personal vector in logits $b(x)+A(x)z$; matrices are evaluated with linear solves.

The ridge adapter uses $r$ learned features plus an intercept, $\phi(x)\in\mathbb R^{r+1}$:
\begin{align}
C_t&=\lambda I+\sum_j\alpha_{tj}\phi_j\phi_j^\top,\\
W_{tj}&=s\alpha_{tj}(\phi_t^\top C_t^{-1}\phi_j)I_K,
\end{align}
where $s=\exp(\operatorname{clip}(\gamma,-3,3))$ for learned scalar $\gamma$. Both families retain Eq.~\eqref{eq:affine}; the reference damping is $\lambda=1$. Point logits use only the observed residual terms. They equal the mean completed logits if hidden responses have marginals $p_j$; their softmax need not equal the mean completed probability vector.

Matched frequency uses the same encoder and head architectures, separately fitted without a completion penalty. Its logits are $b_t+Vf_O+(n/H)v$ for learned $V,v$, observed count $n$, and frequency vector $f_O$ (zero at $n=0$). Completed logits are $b_t+v+\sum_j(V/H)e_{y_j}$. Equation~\eqref{eq:bound} applies with $W_{tj}=V/H$ and $c_t=b_t+v+\sum_{j\in O}(V/H)e_{y_j}$. The audit concerns the completed class; changing $n$ can change intermediate predictions. Equal widths use random ordering.

Identical frequency maps discard item identity and can yield high coverage with weak personalization.

\subsection{Exact completion certificate}
Write
\begin{equation}
c_t=b_t-\sum_jW_{tj}p_j+\sum_{j\in O}W_{tj}e_{y_j}.
\end{equation}
For candidate class $a$ and competitor $b$, the smallest completed logit margin is
\begin{equation}
L_{tab}=c_{ta}-c_{tb}+\sum_{j\in U}\min_k\{(W_{tj})_{ak}-(W_{tj})_{bk}\}. \label{eq:bound}
\end{equation}

\begin{proposition}
For fixed parameters, cases, and routing based only on cues, Eq.~\eqref{eq:bound} is the attained minimum over all response completions. Class $a$ is the unique winner under every completion if and only if $L_{tab}>0$ for all $b\ne a$. Revealing a response cannot decrease any pairwise lower bound.
\end{proposition}
\begin{proof}
Each hidden response contributes a separate summand. Choosing its minimizing category attains the joint minimum because the completion set is a Cartesian product. Positivity of every pairwise minimum is equivalent to a common unique winner. A reveal restricts the completion set, so its minimum cannot decrease.
\end{proof}
A negative bound supplies a counterexample: assigning each hidden response its minimizing category makes the competitor beat class $a$.

\paragraph{Auditing Observed Fisher} This predictor recomputes its Fisher update from the observed ratings and their cues. Its partial logits are
\begin{align}
D_O&=\lambda I+\sum_{j\in O}A_j^\top S_jA_j,\\
\hat\ell_t^O&=b_t+A_tD_O^{-1}\sum_{j\in O}A_j^\top(e_{y_j}-p_j).
\end{align}
At full history, $D_J$ depends only on cues, giving Eq.~\eqref{eq:affine} with uniform weights. We apply the completed-map bounds to the unchanged partial winner. Passing certifies agreement with every completed history, without retraining. Intermediate updates may disagree because their curvature changes. A stopping rule must therefore retain the first certified class rather than continue updating it.

The guarantee conditions on a fitted adapter and completion set. For a specified finite family of fits $\mathcal M$, a common class is certified by $\min_{m\in\mathcal M}L^{(m)}_{tab}>0$ for every competitor. This covers the listed fits, not all plausible parameters, and is stronger than certifying their mean logits. At evaluation, maps and margins use IEEE double precision with threshold $10^{-8}$ in original logit units. Certification requires every competing margin to exceed this threshold; smaller positive margins remain unresolved. Point ties use the first category index. Numerical checks are not interval arithmetic verification.

\subsection{Supporting response acquisition}
For response $j$ and class pair $(a,b)$, define its removable interval width
\begin{align}
d_{tjabk}&=(W_{tj})_{ak}-(W_{tj})_{bk},\\
\Delta_{tj}^{ab}&=\max_k d_{tjabk}-\min_k d_{tjabk}.
\end{align}
We sum $\Delta_{tj}^{ab}$ over all targets and ordered class pairs, then rank responses once. Reveals do not change this order. At any budget, the largest scores minimize remaining total width because the objective separates by position. This is width optimality, not optimality for certification count or loss.

Expected gain rescans hypothetical reveals, weighting newly certified targets by $p_{jk}=\operatorname{softmax}(b_j)_k$. Section~\ref{sec:extensions} tests calibrated and personal probabilities. Scores within $10^{-12}$ of the maximum use width order to break ties. Hidden labels never score test queries. This myopic objective optimizes immediate expected count.

\paragraph{Computational cost} After encoding and routing, Fisher needs $O(HKr^2)$ shared cue curvature work and $O(Hr^2+r^3+Kr^2+HK^2r)$ per target to form its maps. Ridge costs $O(Hd^2+d^3+HK^2)$ per target, with $d=r+1$. Target costs multiply by $T$. Dense margins and width scores each take $O(THK^3)$; ranking adds $O(H\log H)$. The implementation stores $O(THK^2)$ maps and materializes $O(THK^3)$ pair differences. At budget $B$, lookahead repeats $O(THK^3)$ work per retrieval. The dense evaluator does not exploit ridge's scalar identity structure.

\subsection{Training}
The encoder has 24 tanh units and rank four. We average full and masked history cross entropy, with training budgets zero, one quarter, one half, or full history. Fisher and ridge add $\beta\log\sum_b\exp(-L_{ta_tb})$ over at most 12 sampled targets, with $\beta=.03$ and detached full history winner $a_t$. Adam uses learning rate $.01$, weight decay $.001$, gradient clipping at 10, and 60 epochs across cohorts. Layperson sensitivity varies rank (2, 4, 8), width (12, 24, 48), damping (.25, 1, 4), and penalty (0, .03, .06) one factor at a time, without replacing the reference model.

\paragraph{Response influence control} A separate experiment freezes an unpenalized ridge teacher and multiplies each map by a cue-only gate in $[f,1.5]$, initialized at one. Floors $f=0,.5$ are crossed with penalties $\beta=0,.03$. The $.5$ floor preserves half of each single-response logit contrast, not class changes or accuracy. 

\section{Experimental Design}
All experiments retrospectively hide recorded responses.
\subsection{Cohorts}

\textbf{RepVig}: 198 laypeople each rate 20 of 45 vignettes as emergency, nonemergency, or self care, giving 3,960 judgments. A hash divides each history into ten calibration and ten target responses. The study reports informed consent, TU Berlin approval AWB\_KOP\_2\_230711, and registration DRKS00032895 \cite{repvig,repvigdata}.

\textbf{Schmieding}: 91 US Mechanical Turk participants each rate 45 vignettes at the same urgency levels, giving 4,095 judgments. A hash divides histories into 22 calibration and 23 target responses. The study reports informed consent and TU Berlin approval FEU\_03\_20180615 \cite{schmieding}. Both cohorts use one-hot vignette identity.

\textbf{Psychopathology assessment}: 108 practicing, mostly early career clinicians provide 132 sheets across three simulated interviews \cite{lenz,lenzdata}. Each sheet rates 100 Association for Methodology and Documentation in Psychiatry (AMDP) items, reduced to absent, present, or not assessable. The last is an observed category, not a mask. A hash assigns 50 calibration and 50 target items. Features identify item and interview; expert ratings serve only secondary evaluation. Grouped folds test new clinicians on familiar items and interviews. The study reports written consent and Heidelberg approval 2024-567 \cite{lenz}.

\textbf{AnnoMI}: the full public release contains 133 motivational interviewing demonstrations annotated by ten experts \cite{annomi}. We use 64 therapist utterances selected by hash from seven jointly annotated interviews as calibration. The other 126 interviews, from 112 source videos, supply 4,666 target ratings of therapist behavior in four categories. Five expert folds under three new seeds hold out experts and exclude their target source videos from fitting. TF IDF and 32 SVD features use training target texts and shared calibration texts only; no target interview identity is supplied. AnnoMI was collected independently and contains demonstrations rather than patient encounters. Each target interview has one annotator, so expert and interview assignments are confounded.

\subsection{Splits, models, and metrics}
RepVig, Schmieding, and psychopathology use three participant folds under three seeds, grouping clinicians across sheets and excluding test participants from training. Controls are population and ordinary frequency MLPs, matched frequency, Observed Fisher, and target conditioned attention pooling encoded case/response pairs. The reference tables do not audit Observed Fisher; a separate checkpoint audit below does. Ordinary frequency receives no certificate. Population ignores history and is trivially invariant.

Budgets are 0, 25, 50, 75, and 100\% of history, rounded down. Outcomes are log loss in nats, accuracy, coverage, certified error, and Brier score. Tables, figures, and contrasts average targets within sheet, sheets within person, and seeds within person, then weight people equally. Paired 95\% intervals use 10,000 person bootstrap draws, keeping sheets together. These exploratory intervals condition on fitted models, exclude fitting and architecture search uncertainty, and are unadjusted unless stated.

Selections are formed within each fitted seed. A target is history-active when its predicted class differs between zero and full history. Active coverage is the certified fraction of this subset. Conditional metrics divide equally person weighted joint mass by selected mass; empty subsets are undefined. Confidence is the largest softmax probability from the partial history logits. It ranks individual targets, with fractional selection at the boundary to match certificate coverage exactly; no deployment threshold is selected. Top label calibration error uses ten equal width bins and the same weights, then averages seeds.

Internal protocols preceded computations without external preregistration; clinician settings preceded inspection of its results. 

\subsection{Additional controls}
Exploratory clinician controls select width (12, 24, 48) and epochs (40, 80, 120) using 20\% of training clinicians and mean half/full validation loss, then refit on all training clinicians. A stricter test crosses clinician partitions with held psychiatric interviews, excludes test clinicians and interviews from fitting, and uses item identity alone. Half of the new interview's ratings supply calibration. These three interviews were previously inspected.

Confidence targets are reselected within each bootstrap draw at matched weighted coverage. Category metrics use the same person weights before conditioning. Joint selection ranks confidence within certified predictions. Class weighting uses inverse training category frequencies, with unchanged architecture, epochs, and masks; it tests equal category emphasis, not clinical error costs. 

\subsection{Learned policies, calibration, and fitted model variation}
\label{sec:extensions}
This extension reserves disjoint people for fitting, policy selection, temperature calibration, and testing. Selection and calibration each receive 15\% of outer training people, rounded up; no refitting follows selection. Its smaller fitting sets require separate comparisons.

Seven policies share one frozen ridge: random, width, three expected gain variants, learned immediate loss reduction, and terminal loss. Gain variants use raw, temperature-scaled, or updated personal probabilities. Our control regresses realized loss reductions for a frozen predictor, motivated by DIME \cite{dime}, without estimating mutual information or inheriting its guarantees. A second policy uses REINFORCE on final half-budget loss \cite{reinforce}. Adam at $.003$ trains for 300 epochs, with selection at epochs 100, 200, or 300. Hidden labels enter training objectives, never test query scores.

We use temperature scaling \cite{temperature}, fitting temperatures in $[.25,4]$ on calibration people; classes and certificates are unchanged. Five ridge initializations also share random masks and a width order based on their summed scores. These fits share one original three-fold partition. 

\subsection{Response preservation experiment}
Nine clinician and 15 AnnoMI splits compare a 60-epoch teacher, four gates trained for 60 additional epochs, and an unconstrained continuation with the same additional budget. AnnoMI controls receive 60 epochs, sampling 128 targets per expert per epoch and evaluating all targets. Three common random half masks test coverage on the same teacher-active subset. AnnoMI averages targets within expert, then experts equally. Intervals resample its ten experts. Source-video exclusions preceded inspection of external performance. Earlier overlapping-source fits are diagnostic only.

\subsection{Completion agreement without refitting}
The exploratory checkpoint audit uses nine tuned clinician and 15 AnnoMI Observed Fisher fits, saved frequency and ridge controls, and three shared random half masks. Clinician masks differ from the tuned table. Fixed width acquisition and an affine partial rule use the same Fisher parameters, isolating the partial update. No models are refitted or selected for coverage.

\subsection{Stopping at the first certificate}
The post-review stopping replay evaluates every prefix under the checkpoint audit's three random orders and fixed width order. It retains each target's first certified class, falling back to the full-history prediction if unresolved. We distinguish responses needed per target from those needed for all targets sharing a history. No models are refitted; intervals use 10,000 person bootstrap draws.

\section{Results}
\subsection{Completion agreement without changing the predictor}
With weights and revealed responses held fixed, replacing Observed Fisher's partial update with the affine rule raises random-mask loss by .0578 nats in clinicians (95\% CI [.0472,.0690]) and .0591 in AnnoMI ([.0459,.0720]). Both rules certify exactly the same targets (Table~\ref{tab:agreement}), preserving the lower-loss predictions without reducing measured coverage.

Against frequency, Observed Fisher's loss differences are $-.1071$ ($[-.1261,-.0867]$) in clinicians and $-.0223$ ($[-.0321,-.0121]$) in AnnoMI. Against ridge, its loss is lower by .0621 ([.0476,.0768]) and .0174 ([.0066,.0273]), respectively. These comparisons share random masks. With width acquisition, the clinician difference from ridge is $-.0139$ ($[-.0295,.0022]$), which includes zero.

\begin{table*}[!t]
\centering\small
\caption{Completion agreement at half history without refitting. Random masks are shared; width masks differ by model. Active coverage uses each model's own active subset. The two Observed Fisher point rules share weights and masks.}
\label{tab:agreement}
\begin{tabular}{llrrrrrr}\toprule
&&\multicolumn{3}{c}{Clinicians: 25/50 ratings}&\multicolumn{3}{c}{AnnoMI: 32/64 ratings}\\
Predictor & Retrieval & Loss &\shortstack{Coverage\\(\%)}&\shortstack{Active\\coverage (\%)}& Loss &\shortstack{Coverage\\(\%)}&\shortstack{Active\\coverage (\%)}\\\midrule
Frequency & Random & 0.6916 & n/a & n/a & 0.8492 & n/a & n/a\\
Ridge & Random & 0.6466 & 20.8 & 3.9 & 0.8442 & 41.4 & 0.0\\
Observed Fisher & Random & 0.5845 & 5.4 & 1.4 & 0.8269 & 10.5 & 0.2\\
Same fit, affine point & Random & 0.6423 & 5.4 & 1.4 & 0.8860 & 10.5 & 0.2\\
Ridge & Width & 0.5890 & 57.9 & 37.0 & 0.8402 & 48.2 & 0.5\\
Observed Fisher & Width & 0.5751 & 13.7 & 8.0 & 0.8286 & 28.6 & 15.8\\
Same fit, affine point & Width & 0.6012 & 13.7 & 8.0 & 0.8498 & 28.6 & 15.8\\
\bottomrule\end{tabular}\end{table*}

Observed Fisher's history-active fractions are 32.1\% in clinicians and 30.9\% in AnnoMI. Width certifies 8.0\% and 15.8\% of those active targets, versus 13.7\% and 28.6\% overall. Error among its width-certified predictions is 10.3\% and 5.6\%. These subsets are not matched for coverage.

\begin{figure*}[!t]
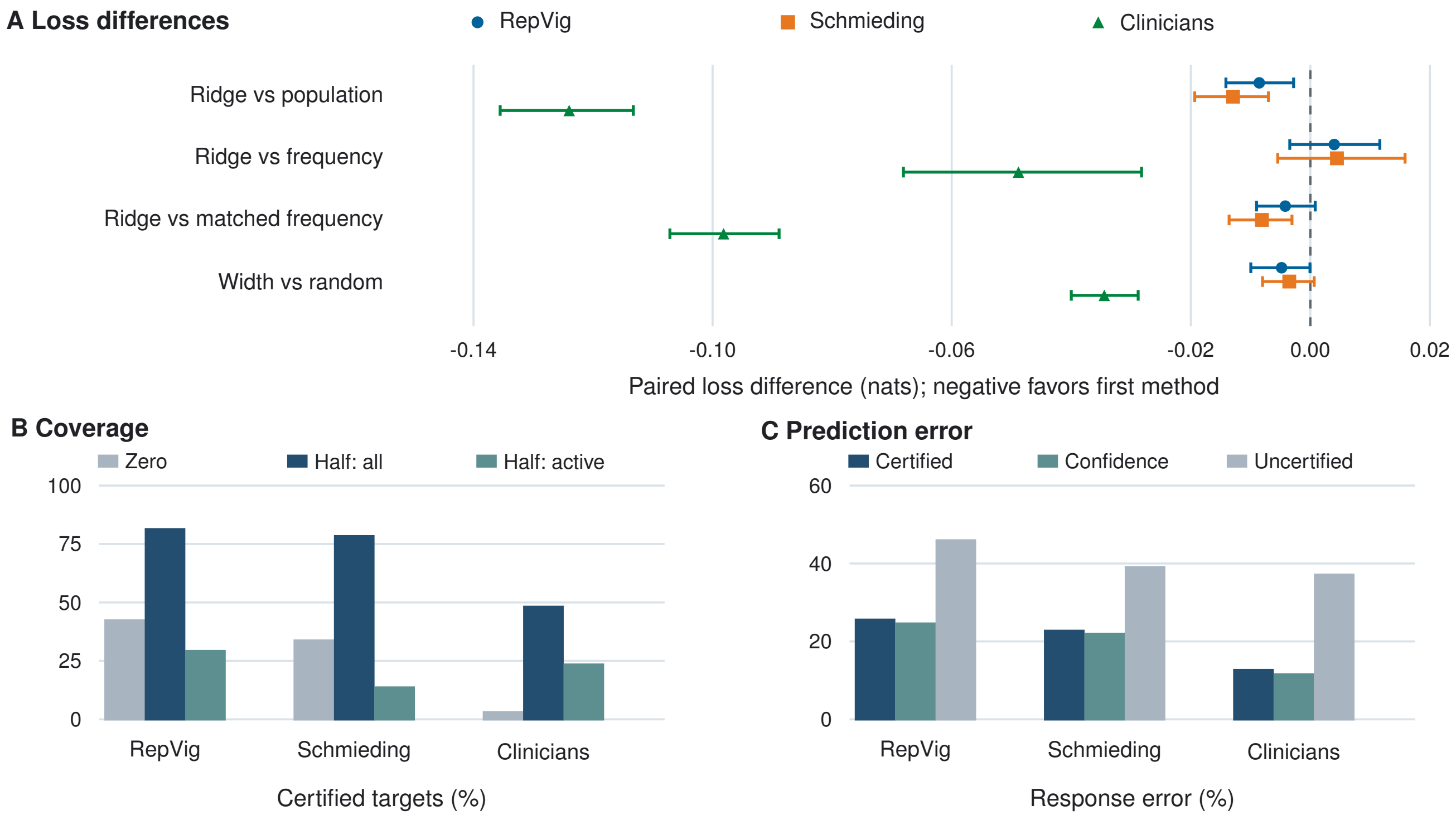

\centering\paperfigure{.98\textwidth}{rq_evidence.pdf}
\caption{Ridge results, equally weighted by person. (A) Paired loss contrasts (95\% intervals); random masks except width versus random. (B) Coverage at zero and half history. (C) Error at half history. Half-history results in B and C use width acquisition; confidence matches certified coverage.}
\label{fig:rqevidence}
\end{figure*}

\subsection{Where predictive gains generalize}
For new clinicians on familiar interviews under shared random masks, ridge minus population loss is $-.1240$ (95\% CI $[-.1355,-.1133]$). Differences from ordinary and matched frequency are $-.0488$ ($[-.0681,-.0282]$) and $-.0982$ ($[-.1071,-.0889]$). Observed Fisher has the lowest loss (.5788; Table~\ref{tab:psych}). These reference masks differ from the checkpoint audit above.

\begin{table}[!htbp]
\caption{Clinician prediction with 25 of 50 ratings. Reference models share random masks; Tuned uses inner validation; Unseen holds out clinicians and an interview. Accuracy and coverage are fractions; n/a: not evaluated.}
\label{tab:psych}\centering\small\setlength{\tabcolsep}{3pt}
\begin{tabular}{lrrrrr}
\toprule &\multicolumn{3}{c}{Reference}&\multicolumn{2}{c}{Loss}\\
Model&Loss&Acc.&Cov.&Tuned&Unseen\\\midrule
Population&.8103&.620&n/a&.7725&1.1174\\
Frequency&.7352&.660&n/a&.6875&1.0177\\
Matched frequency&.7845&.632&.813&.7108&1.0827\\
Attention&.7419&.655&n/a&.6990&n/a\\
Observed Fisher&.5788&.770&n/a&.5823&n/a\\
Global Fisher&.6653&.733&.239&n/a&n/a\\
Routed Fisher&.6666&.731&.248&n/a&n/a\\
Weighted ridge&.6863&.723&.316&.6443&1.0375\\
\bottomrule\end{tabular}\end{table}

With equal validation grids, ridge's loss differences from ordinary and matched frequency are $-0.0433$ (95\% CI $[-0.0583,-0.0271]$) and $-0.0665$ ($[-0.0750,-0.0579]$). This loss advantage is not established on held interviews, where the difference from ordinary frequency is $.0198$ ($[-0.0006,.0446]$). This interval conditions on only three interviews.

\begin{table}[!htbp]
\caption{Prediction at half history. Random masks are shared; Width uses the same ridge model. Loss is in nats; other metrics are fractions. n/a: certificate not evaluated.}
\label{tab:main}\centering\small\setlength{\tabcolsep}{3pt}
\begin{tabular}{llrrrr}\toprule
Model&Policy&Loss&Acc.&Cov.&Cert. error\\
\midrule
\multicolumn{6}{l}{RepVig ($N=198$)}\\
Population&random&.6940&.702&n/a&n/a\\
Frequency&random&.6814&.705&n/a&n/a\\
Matched frequency&random&.6896&.700&.922&.281\\
Attention&random&.6822&.703&n/a&n/a\\
Observed Fisher&random&.6939&.697&n/a&n/a\\
Global Fisher&random&.6890&.705&.659&.236\\
Routed Fisher&random&.6899&.705&.654&.236\\
Weighted ridge&random&.6854&.705&.748&.244\\
Weighted ridge&width&.6806&.706&.813&.256\\
\midrule
\multicolumn{6}{l}{Schmieding ($N=91$)}\\
Population&random&.6464&.728&n/a&n/a\\
Frequency&random&.6290&.733&n/a&n/a\\
Matched frequency&random&.6415&.728&.916&.256\\
Attention&random&.6284&.728&n/a&n/a\\
Observed Fisher&random&.6367&.720&n/a&n/a\\
Global Fisher&random&.6298&.731&.434&.190\\
Routed Fisher&random&.6318&.728&.454&.190\\
Weighted ridge&random&.6334&.732&.730&.212\\
Weighted ridge&width&.6299&.738&.783&.227\\
\bottomrule\end{tabular}
\end{table}

Ridge improves loss over population in both layperson cohorts, but neither comparison with ordinary frequency establishes an advantage (Table~\ref{tab:main}). On AnnoMI, ridge gives .8442 loss versus .8492 for frequency, a difference of $-.0050$ (95\% CI $[-.0133,.0040]$). Population and matched frequency give .8490 and .8513; Observed Fisher gives .8269. Thus the counseling results support Observed Fisher, but not ridge superiority, on unseen experts and source videos. They concern behavior annotations, not diagnoses or treatment decisions.

Ridge improves absent and present recall but reduces not assessable recall (Table~\ref{tab:categories}). Among recorded not assessable targets, its width-certified error is 60.7\% (95\% CI [52.3,68.7]). Counts refer to distinct ratings.

\begin{table}[!htbp]
\caption{Clinician results by recorded category at 25 responses. Frequency and ridge recall use shared random masks; Width uses ridge with width acquisition.}
\label{tab:categories}\centering\small\setlength{\tabcolsep}{3pt}
\begin{tabular}{lrrrrrr}
\toprule & &\multicolumn{3}{c}{Recall (\%)}&\multicolumn{2}{c}{Ridge / width (\%)}\\
Rating & $n$ & Freq. & Ridge & Width & Coverage & Cert. error\\\midrule
Absent & 3748 & 84.5 & 88.8 & 89.1 & 53.2 & 5.1\\
Present & 1962 & 43.6 & 60.8 & 67.3 & 46.1 & 15.5\\
Not assessable & 890 & 37.3 & 26.7 & 28.5 & 30.3 & 60.7\\
\bottomrule\end{tabular}\end{table}

Class reweighting raises not assessable recall from 26.7\% to 63.2\%, but lowers precision from 65.9\% to 38.0\% and raises loss from .6863 to .7861. It also leaves the held-interview gap unresolved. For certified errors, collecting the remaining history cannot correct the completed model's class; improving the predictor or referring the item for human review remains necessary.

\subsection{What certified coverage measures}
Saved minimizing completions overturn both initial predictions in Fig.~\ref{fig:example}.

Matched frequency certifies 92.2\%, 91.6\%, and 81.3\% of half-history predictions in RepVig, Schmieding, and clinicians, despite weaker clinician prediction. In the same cohort order, ridge's zero-response coverage is 42.3\%, 33.7\%, and 3.0\%. Its active fractions are 6.6\%, 4.8\%, and 23.6\%; half-budget width coverage on these active subsets is 29.2\%, 13.6\%, and 23.4\%. High aggregate coverage can therefore coexist with limited response dependence.

With width acquisition, error among ridge-certified targets is 25.6\%, 22.7\%, and 12.6\% across these cohorts (Fig.~\ref{fig:rqevidence}). Confidence selection at matched coverage gives 24.6\%, 21.9\%, and 11.5\%; paired intervals favor confidence in each cohort. Requiring both confidence and certification gives 24.0\% clinician coverage and 7.83\% error. Equal observed error to confidence alone does not establish equivalence (difference interval $[-0.38,0.39]$ percentage points).

In the calibration experiment, temperature scaling reduces clinician width loss to .6540, Brier score from .3734 to .3702, and calibration error from .0425 to .0268, preserving classes and certificates. Intervals for the loss changes include zero in both layperson cohorts.

Under common width masks, average individual coverage is 48.5\%, mean-logit coverage is 49.6\%, and one class is certified across all five fits on 40.5\% of clinician targets. Random masks give 32.2\%, 33.1\%, and 27.0\%. All five partial classes agree on 95.6\% of width targets, but this alone does not imply completion invariance. The common certificate covers only the specified fits.

\subsection{Can the penalty preserve personalization?}
Removing the routed Fisher penalty lowers width loss from .6167 to .5974 but reduces coverage from 47.5\% to 13.8\%. Because the active fractions also change, the gate experiment compares certification on a common subset defined by the unpenalized teacher.

At floor $.5$, the penalty raises aggregate coverage but worsens loss by .0066 nats in clinicians (95\% CI [.0055,.0079]) and .0047 in AnnoMI ([.0022,.0071]). It does not improve certification on the common teacher-active subset under random masks. Unconstrained continuation has lower loss in both cohorts. Preserving individual response contrasts therefore does not establish a better prediction and certification tradeoff.

\subsection{Choosing which responses to retrieve}
For ridge, fixed width acquisition improves clinician loss over random by .0345 nats (95\% CI [.0288,.0400]). The loss difference is $-.00481$ in RepVig ($[-.00996,-.00006]$) and $-.0035$ in Schmieding ($[-.0080,.0006]$). The first narrowly excludes zero among exploratory comparisons; the second does not. Figure~\ref{fig:rqevidence} also shows their coverage differences.

Expected gain raises coverage over width in both layperson cohorts without established loss gains. In clinicians, it instead raises loss by .0207 ([.0157,.0260]) and reduces coverage by 1.47 points ([0.66,2.27]). Calibrated and personal response probabilities do not close this gap.

The learned-policy comparison uses separate selection and calibration people. Here clinician width loss is .6594. Learned loss acquisition improves it by .0050 (95\% CI [.0022,.0078]), while coverage falls from 50.7\% to 49.1\%. Terminal loss optimization gives .6577 and 47.1\% coverage, without an established loss gain. Neither learned policy establishes improvement in the layperson cohorts.

\subsection{Stopping at the first certificate}
With Observed Fisher, width reduces the mean responses needed per target from 38.07 to 34.75 of 50 in clinicians and from 47.36 to 42.11 of 64 in AnnoMI (Table~\ref{tab:stopping}). Paired reductions are 3.32 (95\% CI [3.18,3.45]) and 5.25 ([4.69,5.93]). However, settling every target needs nearly the full shared history. All retained predictions agree with the actual completed model; their errors remain 22.1\% and 32.2\%. This evaluates model-level stopping, not clinical workload reduction. The stopping replay evaluated 270,384 records.

\begin{table}[!htbp]
\centering\small
\caption{Mean ratings needed to retain Observed Fisher predictions at the first certificate. The all-target count is the latest stopping point among targets sharing a history.}
\label{tab:stopping}
\begin{tabular}{llrr}\toprule
Cohort & Order & Per target & All targets\\\midrule
Clinicians ($H=50$) & Random & 38.07 & 49.62\\
 & Width & 34.75 & 49.23\\
AnnoMI ($H=64$) & Random & 47.36 & 64.00\\
 & Width & 42.11 & 64.00\\\bottomrule
\end{tabular}
\end{table}

The 72-checkpoint audit (506,970 records) found no certified disagreement with completed predictions. Independent NumPy checks reproduced sampled margins within $2.23\times10^{-15}$, and small-history enumeration checked the bounds and gain scores.

\FloatBarrier
\section{Discussion}
The main methodological finding is that completion auditing can preserve a useful partial prediction rule (RQ1). Observed Fisher and its affine partial control share fitted parameters and certify exactly the same evaluated targets, yet the observed-curvature update predicts better in both audits. This supports separating the choice of partial predictor from the calculation of completion bounds. The completed model must still retain the response structure needed by the bound. Low coverage measures how rarely completion agreement can be certified under unrestricted assignments; it does not measure clinical safety or the probability of a revision.

The predictive evidence depends on the task and comparison (RQ2). Ridge improves familiar-interview prediction but does not establish superiority over frequency on held psychiatric interviews or AnnoMI. Under shared random masks, Observed Fisher has lower loss in both checkpoint audits but certifies fewer targets than ridge. A choice based only on loss would therefore differ from one based only on coverage. The not assessable category and the unsuccessful gate experiment further show why neither aggregate loss nor aggregate coverage is sufficient.

The supporting acquisition analysis exposes the same tension (RQ3): width improves clinician prediction over random, while learned loss acquisition yields a further gain at lower coverage. Full EDDI, DIME, and other reinforcement learning systems remain untested. Confidence selection gives lower error than certification at matched coverage. The faithful counterfactual audit answers whether unavailable history can overturn the current class in the completed model. This describes the fitted predictor, not the causal or clinical reasons for a rating \cite{rudin,ghassemi}. It neither justifies stopping clinical assessment nor guarantees unchanged intermediate updates.

That auditing role still needs prospective evaluation. The present experiments use retrospective masks, known target cues, and equal query costs. Three simulated psychiatric interviews and counseling demonstrations do not establish performance in patient encounters. AnnoMI broadens source coverage, but only ten experts contribute labels and their interview assignments are confounded. New clinicians and interviews, predictions committed before outcome collection, and measured response burden are needed to assess clinical usefulness.

\section{Conclusion}
Completion auditing checks whether missing ratings can overturn the current class in a fixed completed model. Exact bounds for affine response maps leave the partial predictor unchanged. In the clinician and counseling audits, Observed Fisher had lower loss than frequency and its affine partial control, while certifying the same evaluated targets as the latter. Across four datasets, predictive gains and coverage varied by task. Width acquisition increased coverage and reduced the ratings needed per target, but settling all targets required nearly the full shared history. Confidence selection gave lower error at matched coverage for ridge in the three reference cohorts. Certification establishes completion agreement but can preserve incorrect predictions. Prospective studies must establish whether this audit improves a clinical task or reduces response burden before it guides assessment stopping.


\begin{thebibliography}{99}
\bibitem{lenz} E. Lenz, J. Naamanka, W. Trabert, R. Bottlender, B. Malchow, A. Meyer-Lindenberg, T. Gradinger, and E. Schwarz, ``Benchmarking large language models against practicing clinicians on psychopathological assessment,'' \emph{npj Digital Medicine}, vol. 9, art. 518, 2026, doi:10.1038/s41746-026-02852-7.
\bibitem{multimodalreview} M. A. Raju, P. Siddappa, M. S. H. Al Amin, and R. Ma, ``Multimodal machine learning in healthcare: A tutorial and review,'' \emph{Preprints}, 2026, doi:10.20944/preprints202512.1445.v2.
\bibitem{chexfuse} M. A. Raju, P. Siddappa, M. S. H. Al Amin, R. Ma, and M. R. Ahammed, ``CheXFuse: Guided differential joint attention fusion for robust multimodal chest X-ray classification in data-scarce medical settings,'' arXiv:2610.25617, 2026.
\bibitem{cirrguide} M. A. Raju and R. Ma, ``CirrGuide: A deep cascaded framework for liver cirrhosis segmentation and severity classification from T2-weighted MRI,'' arXiv:2609.14010, 2026.
\bibitem{multiplicity} A. P. Meyer, A. Albarghouthi, and L. D'Antoni, ``The dataset multiplicity problem: How unreliable data impacts predictions,'' in \emph{FAccT}, 2023, doi:10.1145/3593013.3593988.
\bibitem{labelcert} M. Sabanayagam et al., ``Exact certification of (graph) neural networks against label poisoning,'' in \emph{ICLR}, 2025.
\bibitem{nbmissing} S. Bian, X. Ouyang, Z. Fan, and P. Koutris, ``Naive Bayes classifiers over missing data: Decision and poisoning,'' in \emph{ICML}, PMLR 235, 2024, pp. 3913--3934.
\bibitem{knnrepairs} A. Z. Fan and P. Koutris, ``Certifiable robustness for nearest neighbor classifiers,'' in \emph{ICDT}, LIPIcs 220, 2022, pp. 6:1--6:20, doi:10.4230/LIPIcs.ICDT.2022.6.
\bibitem{alertconcordance} P. Siddappa, M. A. Raju, S. H. Al Amin, and R. Ma, ``Alert-list concordance for evaluating imputation strategies in ICU sepsis prediction,'' arXiv:2610.22635, 2026.
\bibitem{r2d2} L. Bertinetto, J. F. Henriques, P. H. S. Torr, and A. Vedaldi, ``Meta-learning with differentiable closed-form solvers,'' in \emph{ICLR}, 2019.
\bibitem{ttr} K. A. Wang, J. Shi, and E. B. Fox, ``Test-time regression: A unifying framework for designing sequence models with associative memory,'' \emph{J. Mach. Learn. Res.}, vol. 27, no. 169, pp. 1--41, 2026.
\bibitem{faast} G. Bao et al., ``FAAST: Forward-only associative learning via closed-form fast weights for test-time supervised adaptation,'' arXiv:2605.04651v2, 2026.
\bibitem{alpaca} J. Harrison, A. Sharma, and M. Pavone, ``Meta-learning priors for efficient online Bayesian regression,'' in \emph{Algorithmic Foundations of Robotics XIII}, Springer Proc. Adv. Robot., vol. 14, 2020, pp. 318--337, doi:10.1007/978-3-030-44051-0\_19.
\bibitem{anp} H. Kim et al., ``Attentive neural processes,'' in \emph{ICLR}, 2019.
\bibitem{selective} R. El-Yaniv and Y. Wiener, ``On the foundations of noise-free selective classification,'' \emph{J. Mach. Learn. Res.}, vol. 11, pp. 1605--1641, 2010.
\bibitem{selectivednn} Y. Geifman and R. El-Yaniv, ``Selective classification for deep neural networks,'' in \emph{NeurIPS}, 2017.
\bibitem{conformalmissing} M. Zaffran, A. Dieuleveut, J. Josse, and Y. Romano, ``Conformal prediction with missing values,'' in \emph{ICML}, PMLR 202, 2023, pp. 40578--40604.
\bibitem{cat} D. J. Weiss and G. G. Kingsbury, ``Application of computerized adaptive testing to educational problems,'' \emph{J. Educ. Meas.}, vol. 21, no. 4, pp. 361--375, 1984, doi:10.1111/j.1745-3984.1984.tb01040.x.
\bibitem{catstop} X. Luo, D. Kim, and P. Dickison, ``Projection-based stopping rules for computerized adaptive testing in licensure testing,'' \emph{Appl. Psychol. Meas.}, vol. 42, no. 4, pp. 275--290, 2018, doi:10.1177/0146621617726790.
\bibitem{eddi} C. Ma, S. Tschiatschek, K. Palla, J. M. Hern\'andez-Lobato, S. Nowozin, and C. Zhang, ``EDDI: Efficient dynamic discovery of high-value information with partial VAE,'' in \emph{ICML}, PMLR 97, 2019, pp. 4234--4243.
\bibitem{dime} S. Gadgil, I. Covert, and S.-I. Lee, ``Estimating conditional mutual information for dynamic feature selection,'' in \emph{ICLR}, 2024.
\bibitem{brig} J. Feng, Q. Li, and Y. Li, ``BRiG-AFA: Bellman risk-to-go learning for non-myopic active feature acquisition,'' arXiv:2608.02305, 2026.
\bibitem{afabench} V. Sch\"utz, H. Wu, R. Rezvan, L. Aronsson, and M. Haghir Chehreghani, ``AFABench: A generic framework for benchmarking active feature acquisition,'' in \emph{ACM SIGKDD}, 2026, pp. 9765--9776, doi:10.1145/3770855.3817493.
\bibitem{repvig} M. Kopka et al., ``The RepVig framework for designing use-case specific representative vignettes and evaluating triage accuracy of laypeople and symptom assessment applications,'' \emph{Scientific Reports}, vol. 14, art. 30614, 2024, doi:10.1038/s41598-024-83844-z.
\bibitem{repvigdata} M. Kopka et al., ``Data set supplementing `A Use-Case Specific Framework for Designing Representative Vignettes (RepVig) and Evaluating Triage Accuracy of Laypeople and Symptom-Assessment Applications','' Zenodo, 2024, doi:10.5281/zenodo.12805049.
\bibitem{schmieding} M. L. Schmieding et al., ``Benchmarking triage capability of symptom checkers against that of medical laypersons: Survey study,'' \emph{J. Med. Internet Res.}, vol. 23, no. 3, e24475, 2021, doi:10.2196/24475.
\bibitem{lenzdata} E. Lenz et al., ``LLMs\_psychopathologies,'' public data and code, 2026. Accessed: Oct. 4, 2026. [Online]. Available: \url{https://github.com/42elenz/LLMs_psychopathologies}
\bibitem{annomi} Z. Wu, S. Balloccu, V. Kumar, R. Helaoui, D. Reforgiato Recupero, and D. Riboni, ``Creation, analysis and evaluation of AnnoMI, a dataset of expert-annotated counselling dialogues,'' \emph{Future Internet}, vol. 15, no. 3, art. 110, 2023, doi:10.3390/fi15030110.
\bibitem{reinforce} R. J. Williams, ``Simple statistical gradient-following algorithms for connectionist reinforcement learning,'' \emph{Machine Learning}, vol. 8, pp. 229--256, 1992, doi:10.1007/BF00992696.
\bibitem{temperature} C. Guo, G. Pleiss, Y. Sun, and K. Q. Weinberger, ``On calibration of modern neural networks,'' in \emph{ICML}, PMLR 70, 2017, pp. 1321--1330.
\bibitem{rudin} C. Rudin, ``Stop explaining black box machine learning models for high stakes decisions and use interpretable models instead,'' \emph{Nature Machine Intelligence}, vol. 1, pp. 206--215, 2019, doi:10.1038/s42256-019-0048-x.
\bibitem{ghassemi} M. Ghassemi, L. Oakden-Rayner, and A. L. Beam, ``The false hope of current approaches to explainable artificial intelligence in health care,'' \emph{The Lancet Digital Health}, vol. 3, no. 11, pp. e745--e750, 2021, doi:10.1016/S2589-7500(21)00208-9.
\end{thebibliography}
\end{document}